\documentclass{article}

\PassOptionsToPackage{numbers}{natbib}
\usepackage[final]{neurips_wrl2021}

\usepackage[utf8]{inputenc} % allow utf-8 input
\usepackage[T1]{fontenc}    % use 8-bit T1 fonts
\usepackage{hyperref}       % hyperlinks
\usepackage{url}            % simple URL typesetting
\usepackage{booktabs}       % professional-quality tables
\usepackage{amsfonts}       % blackboard math symbols
\usepackage{nicefrac}       % compact symbols for 1/2, etc.
\usepackage{microtype}      % microtypography
\usepackage{amsmath}
\usepackage{amssymb}
\usepackage{bm}
\usepackage{bbm}
\usepackage{algorithm}
\usepackage{algorithmic}
\usepackage{graphicx}
\usepackage{amsfonts}
\usepackage{booktabs}
\usepackage{adjustbox}
\usepackage{siunitx}
\usepackage[font=small]{caption}

\usepackage{xcolor}

\title{IL-flOw: Imitation Learning from Observation using Normalizing Flows}

\author{%
  Wei-Di Chang, Juan Camilo Gamboa Higuera, Scott Fujimoto, David Meger,  Gregory Dudek \\
  McGill University\\
  Contact: \texttt{wei-di.chang@mail.mcgill.ca} \\
}

\begin{document}
\vspace{-20pt}
\maketitle

\begin{abstract}
  We present an algorithm for Inverse Reinforcement Learning (IRL) from expert state observations only. Our approach decouples reward modelling from policy learning, unlike state-of-the-art adversarial methods which require updating the reward model during policy search and are known to be unstable and difficult to optimize. %These adversarial methods are known to be unstable and difficult to optimize. Furthermore, these methods require action supervision. 
  Our method, IL-flOw, recovers the expert policy by modelling state-state transitions, by generating rewards using deep density estimators trained on the demonstration trajectories, avoiding the instability issues of adversarial methods. We demonstrate that using the state transition log-probability density as a reward signal for forward reinforcement learning translates to matching the trajectory distribution of the expert demonstrations, and experimentally show good recovery of the true reward signal as well as state of the art results for imitation from observation on locomotion and robotic continuous control tasks.
\end{abstract}

\section{Introduction}
Imitation learning (IL) is an effective class of algorithms for designing and optimizing controllers for robot systems. While recent advances in Reinforcement Learning have shown it is capable of producing agents that learn robot controllers from scratch, IL remains a more practical alternative for cases where it is easier to specify robot behaviours through examples than through rewards. We take an approach similar to existing work \cite{ziebart2008maximum, ho2016generative} on learning from demonstrations (LfD). We use expert data to build a reward model to be maximized with existing RL algorithms.  Unlike LfD, where an expert demonstrates which \emph{actions} to perform at some robot states, we focus on the case where action supervision is not available: the agent only gets access to a dataset of state/observations sequences -- a setting known as Imitation Learning from Observations (ILfO). While LfD usually requires providing demonstrations by teleoperation of the robot, ILfO aims to utilize streams of state and observational data, much like a human can learn to do a task by watching other people. %Learning from visual demonstrations, also known as Learning from Observation (LfO) has garnered much attention in recent years, 

%[more intro text]

We formulate the problem of learning from observations as a distribution matching problem: we want to find the policy parameters that result in observation sequences that are similar to those in a dataset of expert demonstrations. This is similar to recent work \cite{ho2016generative, ghasemipour2020divergence, firl2020corl} that uses adversarial optimization. Our approach differs in that we fit a density model on expert observation sequences, which we then use to produce rewards for policy search with RL optimizers, decoupling the policy optimization from the reward learning processes. %Adversarial formulations \cite{ho2016generative, henderson2018optiongan, kostrikov2018addressing} have also been found to generally underperform non-adversarial imitation learning algorithms \cite{arulkumaran2021pragmatic}. 
To the best of our knowledge the closest approaches to ours are \cite{neuraldensityimitation} where they also use neural density estimators although for occupancy measure estimation and address state-action imitation instead of state-only. Their formulation requires a discounted infinite horizon agent however, as opposed to our undiscounted finite horizon RL optimization. And \cite{dadashi2020primal}, which also addresses the lack of smoothness of the KL-divergence objectives, by opting to minimize Wasserstein distance however instead of noise-expanded distributions, although unlike ours their reward signal is non-stationary.
Our imitation signal is non-adversarial, stationary and reusable for downstream tasks.

%Starting from a KL divergence minimization formulation for trajectory matching, we derive an objective suitable for imitation learning from observation using a Reinforcement Learning agent. We leverage density estimation capabilities of normalizing flows to implement this objective and benchmark our approach on high dimensional mujoco continuous control tasks as well as robot manipulation tasks.

% Want contributions here
% Formulation [including undiscounted]
% Noise conditioned nf

\vspace{-5pt}
\section{Background}
% Setting
We formulate our task by an MDP $(\mathcal{S}, \mathcal{A}, \mathcal{P}, r, p_0)$ where $\mathcal{S}$ is the state space, $\mathcal{A}$ is the action space, $\mathcal{P}: \mathcal{S} \times \mathcal{A} \times \mathcal{S} \rightarrow [0,1]$ is the transition dynamics, $r: \mathcal{S} \times \mathcal{A} \rightarrow \mathbb{R}$ is the reward function, $p_0: \mathcal{S} \rightarrow [0,1]$ is the initial state distribution. %, and $\y \in [0,1)$ is the discount factor. 
We define a policy~$\pi: \mathcal{S} \times \mathcal{A} \rightarrow [0,1]$. The probability of a trajectory $\tau = \{s_{0:T}, a_{0:T}\}$ of $T+1$ states and actions when following the policy $\pi$ is given by: $p_\pi(s_{0:T}, a_{0:T}) = p_0(s_0) \prod_{t=0}^{T-1} p(s_{t+1}|s_t, a_t) \pi(a_t|s_t)$. If we are interested in state-only trajectories, then we must consider transitions over states, with the effects of the policy integrated out: $p_\pi(s_{t+1}|s_t) = \int p(s_{t+1}|s_t, a)  \pi(a|s_t) \mathrm{d}a$. It follows that the two quantities of interest are the probability of a trajectory over states 
\begin{align}
    &\text{given a policy }\pi: p_\pi(s_{0:T-1}) = p_0(s_0) \textstyle \prod_{t=0}^{T-1} p_\pi(s_{t+1}|s_t), \\
   &\text{and given an expert }E: p_E(s_{0:T-1}) = p_0(s_0) \textstyle \prod_{t=0}^{T-1} p_E(s_{t+1}|s_t),
\end{align}
where state imitation learning from observations occurs by distribution fitting $p_\pi$ to match $p_E$.
%given $\pi$: $p_\pi(s_{0:T}) = p_0(s_0) \prod_{t=0}^{T-1} p_\pi(s_{t+1}|s_t)$ and given an expert $E$: $p_E(s_{0:T}) = p_0(s_0) \prod_{t=0}^{T-1} p_E(s_{t+1}|s_t)$. 
%\footnote{We might want to add a trajectory termination event, otherwise longer trajectories will become less likely than shorter ones. For some of the results below we remove the dependency on $T$, so it might not be a problem in those cases}
% \begin{equation} \label{eq:traj_lik}    
%     p_\pi(s_{0:T}, a_{0:T}) = p_0(s_0) \prod_{t=0}^{T-1} p(s_{t+1}|s_t, a_t) \pi(a_t|s_t).
% \end{equation}

% If we assume we can only observe expert state sequences then $p_E(s_{0:T}) = p_0(s_0) \prod_{t=0}^{T-1} p_E(s_{t+1}|s_t)$.
% %so we have the following distribution for expert state sequences
% % \begin{align}
% %     p_E(s_{0:T}) = p_0(s_0) \prod_{t=0}^{T-1} p_E(s_{t+1}|s_t).
% % \end{align}
% This means the pr integrate out the effects of the policy, so that we can compare the two distributions:
% \begin{align} \label{eq:states_lik}
%     p_{\pi}(s_{0:T})&=p_0(s_0) \prod_{t=0}^{T-1} \int p(s_{t+1}|s_t, a)  \pi(a|s_t) \mathrm{d}a =p(s_1) \prod_{t=0}^{T-1} p_{\pi}(s_{t+1}|s_t).
% \end{align}
% Given these two definitions, we can state imitation learning from observations as distribution fitting; i.e. finding the policy $\pi$ that causes $p_{\pi}(s_{0:T})$ to match state sequence observations from expert data $p_E(s_{0:T})$.
\vspace{-5pt}
\section{IL-flOw}
% - Intro
%     - Implements RKL ILFO 
%     - We first optimize a NF on the expert dataset
%     - Followed by running forward RL on the learnt density function used as a reward.
% - A. Noise conditioned NF
% - B. Density-based reward function?
% - C. Forward RL on Log Density reward
\vspace{-5pt}
In this section we derive IL-flOw, an Imitation Learning from Observation algorithm that implements trajectory matching by maximizing the log probability of the expert transitions. We begin by reformulating the reverse KL objective as an expression over individual transitions. This suggests a straightforward approach in which we use an approximation of the log probability of expert transitions as a reward signal alongside entropy maximization. Secondly, we present our noise regularization approach for density estimation of expert transitions using normalizing flows. 

%Name-of-alg operates in two phases: First, the density estimator for $p_E(s' | s)$ is trained to convergence by maximum likelihood, after which we train an RL agent on the learnt transition density model, using log probabilities as reward $r(s, s') = \log p_E(s' | s)$, as described above.
\vspace{-5pt}
\subsection{Imitation Learning via a Trajectory Matching Objective} \label{sec:rev_KL_il}
\vspace{-5pt}
Given our objective is matching the distribution of trajectories $p_\pi$ induced by our current policy $\pi$, and the distribution $p_E$ by some expert $E$, in this section we reformulate the reverse KL (RKL) \footnote{Following convention in imitation learning, the reverse KL is defined as $D_\text{KL}\left(p_{\pi}||p_E\right) $} divergence such that it is more amenable for use in a reinforcement learning context. 

%Now, we try the other way around, the reverse KL divergence. Which means that we are evaluating the objective by rolling out the current policy $\pi$. 
% Consider the reverse KL (RKL) divergence [maybe add note about fkl vs rkl "convention"] between the distribution of trajectories induced by our current policy $\pi$ and the distribution of expert trajectories~$p_E$: 
First, consider the RKL between trajectory distributions: \\ 
\begin{align}
    D_\text{KL}\left(p_{\pi}||p_E\right) 
    % &= -\mathbb{E}_{\tau \sim p_{\pi}(\tau)}\left[\log \frac{p_E(\tau)}{p_{\pi}(\tau)} \right]\label{eq:KL_rev} =-\mathbb{E}_{\tau \sim p_{\pi}(\tau)}\left[\log \frac{\prod_{t=0}^{T-1} p_E(s_{t+1}|s_t)}{\prod_{t=0}^{T-1} p_{\pi}(s_{t+1}|s_t)} \right]\\
    &= - \mathbb{E}_{s_{0:T-1} \sim p_{\pi}}\left[\sum_{t=0}^{T-1} \log p_{E}(s_{t+1}|s_t)\right]  + \mathbb{E}_{s_{0:T-1} \sim p_{\pi}}\left[ \sum_{t=0}^{T-1} \log p_{\pi}(s_{t+1}|s_t) \right]
    %&= -( J(\pi) + \mathcal{H}(p_{\pi}) ). 
\end{align}
% Noting the first term resembles an undiscounted reinforcement learning objective $J(\pi)$, where the reward is $r= \log p_{E}(s_{t+1}|s_t)$, and the second term is simply the entropy over $p_\pi$, %%% Dont love this notation overloading thats happening.

%For evaluating $J(\pi)$, we need a model , which we can interpret as a reward if we use RL optimizers. 

%\footnote{TODO: derive the gradient here and show that it'll be like any policy gradient algorithm, with rewards equal to the log probs}\footnote{When using RL optimizers, we need to be careful about treating $\log p_{E}(s_{t+1}|s_t)$ as rewards because the log probabilities are usually negative. If an environment contains terminal states that are not goal states for the task, a policy can maximize rewards by going directly to any of these terminal states. A way to fix this is to remove terminal states or to assign very low rewards to them. A better way is to expand the state $s_t$ to be the concatenation of state variables and a terminal state flag $\bar{s} = [s_t, d_t]$, where $d_t$ is 1 at terminal states and 0 elsewhere. This should encourage $p_E(\bar{s}_{t+1}| \bar{s}_t)$ to assign very low probabilities to terminal states, unless they are task goals, penalizing policies that reach these states with very low rewards. In this case, care must be taken to avoid numerical errors.}.

The first term is the likelihood of policy samples under the expert distribution. The second term corresponds to the entropy of the state-sequence distribution induced by the policy. Note that, by the law of iterated expectations, the second term can be written as the expectation of the per-timestep transition entropies (full derivation in Appendix \ref{ap:equations}) 
\begin{align}
\label{eq:entropies}
\mathcal{H}(p_{\pi}) =  -\mathbb{E}_{s_{0:T-1} \sim p_{\pi}}\left[\sum_{t=0}^{T-1} \log p_{\pi}(s_{t+1}|s_t)\right] 
%& \quad \vdots \nonumber\\
% &= \sum_{i=0}^{T-1} -\mathbb{E}_{s_{0:T} \sim p_{\pi}(\tau)}\left[\log p_{\pi}(s_{i+1}|s_i)\right] \\
% &= \sum_{i=0}^{T-1} -\mathbb{E}_{s_{0:i+1} \sim p_{\pi}(\tau)}\left[\log p_{\pi}(s_{i+1}|s_i)\right] \\
% &= \sum_{i=0}^{T-1} \mathbb{E}_{s_{0:i} \sim p_{\pi}(\tau)}\left[ \mathcal{H}(s_{i+1}|s_i) \right] \\
= \mathbb{E}_{s_{0:T-1} \sim p_{\pi}}\left[\sum_{t=0}^{T-1} \mathcal{H}\left(p_{\pi}(\cdot|s_{t})\right) \right].
\end{align}
If we assume the dynamics are deterministic and invertible\footnote{Given a pair of states $s_t, s_{t+1}$ we can uniquely determine the action $a_t$ that produced it}, we can simplify the expression further by using the change of variables formula\footnote{$|p(x)\mathrm{d}x| = |p(y)\mathrm{d}y|$ if $y= f(x)$ and $f$ is invertible} to express the state sequence entropy in terms of the policy (full derivation in Appendix \ref{ap:equations})
%In this case, we can replace the entropy of the state sequence distribution $\mathcal{H}(p_{\pi})$ with the entropy of the policy $\mathcal{H}(\pi)$. The justification for this is that if the entropy of $\pi(a|s_t)$ increases, when the dynamics are deterministic and invertible, the entropy of $p(s_{t+1}|s_t, a_t) = \int \delta(s_{t+1} = f(s_t, a_t)) \pi(a|s_t) \mathrm{d}a$ can only increase. 
\begin{align}
\label{eq:approximation}
    \mathcal{H}\left(p_{\pi}(\cdot|s_{t})\right) &= -\int p_{\pi}(s_{t+1}|s_t) \log p_{\pi}(s_{t+1}|s_t) \mathrm{d}s_{t+1} \approx \mathcal{H}\left(\pi(\cdot|s_{t})\right). %+ \mathbb{E}_{a_t \sim \pi}\left[\log \left|\frac{\mathrm{d}s_{t+1}}{\mathrm{d}a_{t}}\right| \right].
\end{align}
%For a non-deterministic environment we increasing the entropy of the policy should increase the entropy of the distribution of the next state. \juan{This sentence is unnecessary since we are considering deterministic environments}
While the assumption of invertible dynamics is restrictive, our experiments show that it is a useful approximation for robotics tasks.
Minimizing the KL divergence objective above is equivalent to maximizing the following objective:
\begin{align}
    J(\pi) + \mathcal{H}(p_{\pi}) 
    %&=  \mathbb{E}_{\tau \sim p_{\pi}(\tau)}\left[\sum_{t=0}^{T-1} \log p_{E}(s_{t+1}|s_t)\right] + \mathbb{E}_{\tau \sim p_{\pi}(\tau)}\left[\sum_{t=0}^{T-1} \mathcal{H}(s_{t+1}|s_t) \right] \nonumber\\
    &\approx  \mathbb{E}_{\tau \sim p_{\pi}(\tau)}\left[\sum_{t=0}^{T-1} \log p_{E}(s_{t+1}|s_t) + \sum_{t=0}^{T-1} \mathcal{H}(\pi(\cdot|s_t)) \right]. \label{eq:KL_rev_combined}
     %+ \sum_{t=0}^{T-1} \mathbb{E}_{a_t \sim \pi}\left[\log  \left|\frac{\mathrm{d}s_{t+1}}{\mathrm{d}a_{t}}\right| \right] \right].
\end{align}
% Given enough expert data, we may train a generative model of the transition log-probabilities $\log p_{E}(s_{t+1}|s_t)$ to evaluate $J(\pi)$. 

%Thus our objective becomes
%the objective that we want to maximize, in order to minimize the reverse KL divergence, is
% \begin{align}
%     D_\text{KL}\left(p_{\pi}(\tau)||p_E(\tau)\right) = -(J(\pi) + \mathcal{H}(p_{\pi})) 
%     %&= \mathbb{E}_{\tau \sim p_{\pi}(\tau)}\left[\sum_{t=0}^{T-1} \log p_{E}(s_{t+1}|s_t)\right]  + \mathbb{E}_{\tau \sim p_{\pi}(\tau)}\left[\sum_{t=0}^{T-1}\mathcal{H}(s_{t+1}|s_t) \right] \nonumber\\
%     = \mathbb{E}_{\tau \sim p_{\pi}(\tau)}\left[\sum_{t=0}^{T-1} \log p_{E}(s_{t+1}|s_t) + \sum_{t=0}^{T-1} \mathcal{H}(s_{t+1}|s_t) \right].
% \end{align}
% It follows that minimizing the RKL divergence is equal to maximizing the objective $J(\pi) + \mathcal{H}(p_\pi)$. 
% The first term, $J(\pi)$ is evaluating whether the samples from rolling out the policy have high probability under the expert distribution model, while $\mathcal{H}(p_{\pi})$ is measuring the entropy of the trajectory distribution obtained by rolling out the policy. 
%We can interpret the first term, $J(\pi)$ as a maximum likelihood loss: we want samples from policy evaluations to have high likelihood under the expert transition model $p_E(s_{t+1}|s_t)$.
This objective can be optimized with RL algorithms by setting rewards to $r_t = \log p_{E}(s_{t+1}|s_t)$ and maximizing the undiscounted return, while penalizing the negative entropy of the policy. This suggests a practical algorithm where we can use a finite horizon variant of Soft Actor-Critic \cite{haarnoja2018soft}, which maximizes a reward signal alongside the entropy of the policy. 
\vspace{-5pt}
% NF details
\subsection{Noise Conditioned Normalizing Flows}
\vspace{-5pt}
Our approach requires fitting a model of $p_E(s_{t+1}|s_t)$, using a dataset of demonstrations $D_E$. We use a normalizing flow model to fit $p_E$, a very powerful and expressive type of density estimator. While well-suited for our purpose, these models are known to overfit with little data, leading to poor out of distribution generalization \cite{nalisnick2019, kirichenko2020}. Since we aim to use this density model as a reward model for an RL optimizer, we also want it to be suitable for policy optimization. This means producing reasonably low probabilities for observations that are far from the expert data and likely encountered during policy optimization, while resulting in a smooth optimization landscape. Given that $D_E$ only covers a subset of all possible behaviours that could be encountered during optimization, we have little control on the predicted log-probability density for non-expert behaviour. This results in a noisy, and possibly biased signal for policy optimization outside the support of the training dataset\footnote{This same issue is encountered in GAN training \cite{goodfellow2014generative, principledgan} and prompted solutions such as label smoothing \cite{salimans2016improved}, and Wasserstein critics \cite{arjovsky2017wasserstein}; too sharp a learning signal leads to poor training signals for the generator. It is also discussed in \cite{chinwei2020pdistill} in the context of probability distillation.}.
 
To address the issues above, we fit a set of noise conditioned distributions, $\tilde{p}_{E}(s_{t+1}|s_{t}, h)$ where $ h \sim \mathrm{Uniform}\left[0, h_{max}\right]$ represents the \emph{noise level} -- the magnitude of zero-mean noise added to the training data. At training time, we draw a noise level $h$ and two zero-mean noise samples\footnote{e.g. Normal or Cauchy distributed} $\bm{\epsilon_s}$ and $\bm{\epsilon_{s'}}$ for each expert transition $(\bm{s_E}, \bm{s_E'})$. We set $\bm{\tilde{s}_E } \leftarrow \bm{s_E }+ \bm{h\epsilon_s}$ and $\bm{\tilde{s}_E'} \leftarrow \bm{s_E'} + \bm{h\epsilon_{s'}}$ and fit our model to maximize $\log p(\bm{\tilde{s}_E'}  | \bm{\tilde{s}_E}, h)$. At test time, with $h=0$ we recover the noise-free fitted distribution $\tilde{p}_{E}$, while with $h=h_{max}$ we get a distribution closer in shape to the distribution of the additive noise, as it then dominates over $p_E$. Any intermediate value for $h$ smoothly interpolates between the two. Since the sampled noise is zero-mean, transitions close to the dataset $D_E$ will have the highest log-probability, irrespective of the noise level, providing a signal with tunable smoothness that is useful for policy search. In Appendix \ref{fig:dims} we show an example of varying the noise level $h$. Noise regularization for density estimators has been studied in \cite{rothfuss2020noise}, and noise-conditioned normalizing flows have previously been applied to 3D data \cite{softflow2020}. Previous work restricts the noise level to $h \approx 0$ at test time however, while we actively use the set of noise levels $h \in [h_{min}, h_{max}]$ to control the smoothness of our optimization objective, giving the agent a usable signal at all times during policy optimization.
% we are the first to use the full set of learnt distributions as a means to building an optimization path.
% \weidi{probably remove "high dimensional", all I mean is >3.
% In addition we are the first to use the "full" distribution instead of just h=0 and throwing away the rest but not sure that is worth stating. Also potentially move to the pseudo related works section.}

% The disturbance distribution expands the demonstration states radially in a hypersphere (equally in each direction), 

%This noise regularization process is a form of data augmentation and a regularization method. The noisy data covers parts of the input space that would otherwise be unseen in training, effectively giving non-zero probability to those samples, smoothing the learnt distributions (see Appendix \ref{fig:dims} for a visualization of the noise regularization process).
% The conditional distributions with more added noise have higher entropy (lower highs, higher lows) [bit handwavy here] than their low noise counterpart, even though all the distributions are learnt by a single model, thus ensuring that the highest likelihood is assigned to the noise free samples. [Potentially find way to prove that the training set ends up with highest log p using noise reg]
In the results below, we chose to use Neural Spline Flows~\cite{durkan2019} for density estimation and Soft Actor-Critic \cite{haarnoja2018soft} as the policy optimizer, but our objective and approach are applicable to any combination of a density estimator and optimizer.
\vspace{-5pt}
\subsection{Soft Actor-Critic in time-limited MDPs and adaptive noise conditioning}
% \scott{Maybe just move all of this to appendix? although the Choosing the noise level is interesting so not sure. (Moved to end of Section 3 for space atm) }
% Not discounting . . 
% We use Soft-Actor-Critic \cite{haarnoja2018soft} as our maximum entropy RL algorithm acting in time-limited  MDPs. 

Operating in a finite horizon setting, we augment the state with a time-to-horizon variable $t_H$, representing the number of timesteps to go in an episode, therefore making the actor and critic networks both time-aware. We also augment the action by one additional dimension representing the noise level $h$ at which to sample our density function, thus letting the agent interact with the entirety of the log probability reward signal. We know however that the highest log density is achieved by the training dataset, at the noise-free level $h_{min} = 0$. The agent should learn to choose a low value of $h$ when close to the expert, while as we move away from the expert support the \textit{appropriate} value of $h$ smoothly increases, expanding the support of the density function.

\vspace{-4pt}
\section{Experiments and Results}
\label{sec:experiments}
\vspace{-5pt}

\begin{figure*}[h!]
\centering
  \includegraphics[width=.8\textwidth]{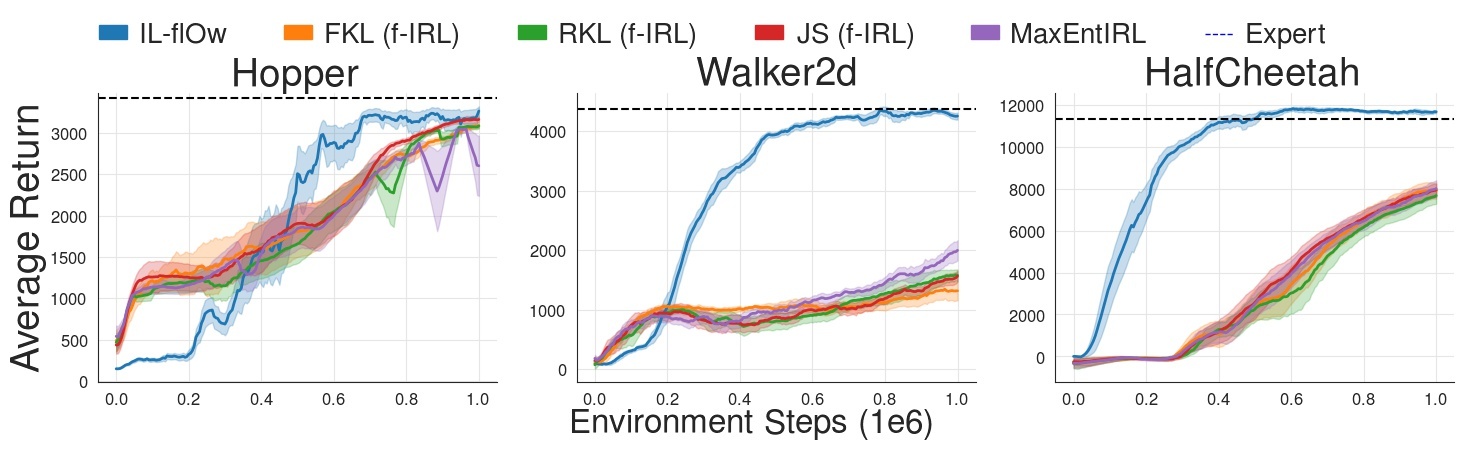}
\vspace{-.5em}
\caption{Learning curves for IL-flOw and 4 other baselines: f-IRL (FKL, RKL, JS), and MaxEntIRL with 40 expert demonstrations across 3 random seeds. The shaded area represents half a standard deviation.}
\label{fig:results}
\vspace{-7pt}
\end{figure*}

We collect $n=150$ demonstration trajectories on three Mujoco \cite{mujoco} simulated environments: Hopper-v2, Walker2d-v2, and HalfCheetah-v2, using a SAC expert trained for 1M timestep, using $n$ random seeds. To evaluate the performance with varying amounts of demonstrations, we use a subset of 40, 20 or 10 trajectories (respectively $(40, 20, 10) \cdot 10^3$ data points) as the training dataset $D_{expert}$ for the density estimator. We compare IL-flOw to the three variants of the f-IRL \cite{firl2020corl} algorithm, as well as a state only version of MaxEnt IRL \cite{ziebart2008maximum}, using the implementations provided by \cite{firl2020corl}. 
MaxEnt IRL  minimizes forward KL divergence in trajectory space under the maximum entropy RL framework.
f-IRL is an imitation learning from observation algorithm that operates by state marginal distribution matching, through optimization of the analytical gradient of any f-divergence (JS, FKL, RKL). It also learns a stationary reward that is reusable, although the imitation agent still faces a moving reward function in training through their iterative training process, while for IL-flOw the reward learning and the RL process are sequential, to convergence. 

\begin{figure*}
\centering
\begin{tabular}{@{}c@{}c@{}c@{}} 
  \includegraphics[width=.32\textwidth]{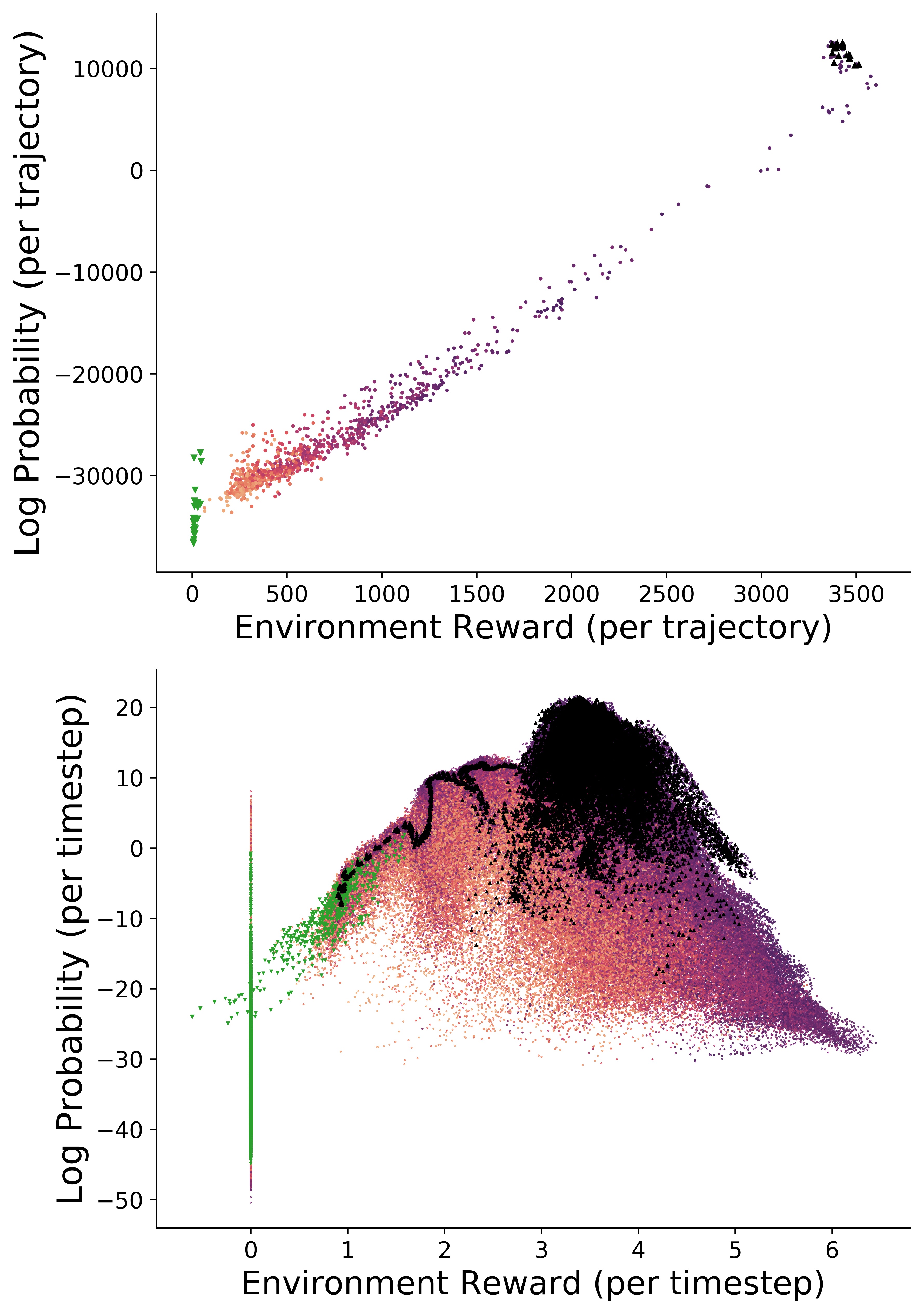}&%{figures/10noises_dim_11_0.01.jpg} &
  \includegraphics[width=.32\textwidth]{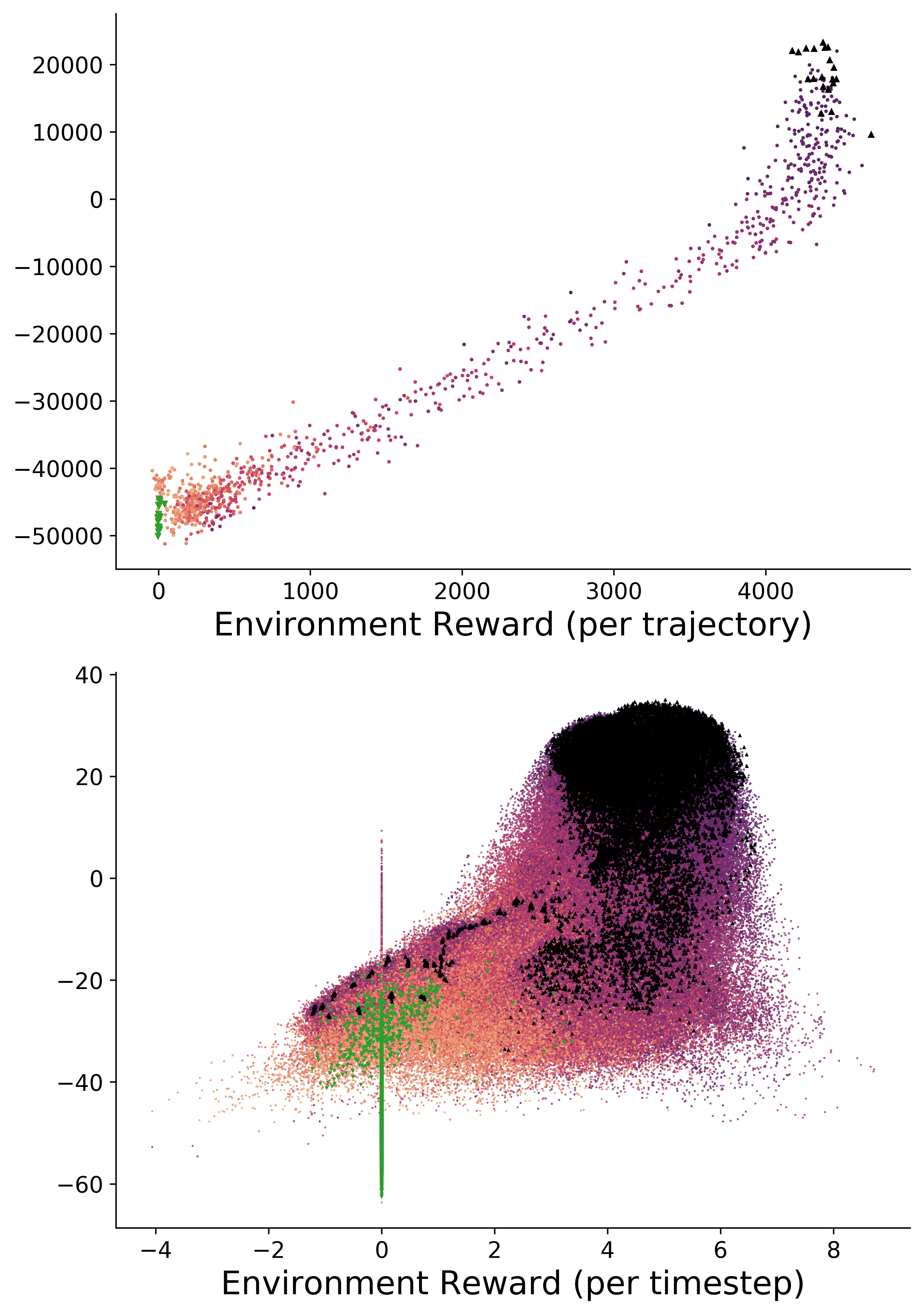}&%{figures/10noises_dim_11_1.505.jpg}  &
  \includegraphics[width=.48\textwidth]{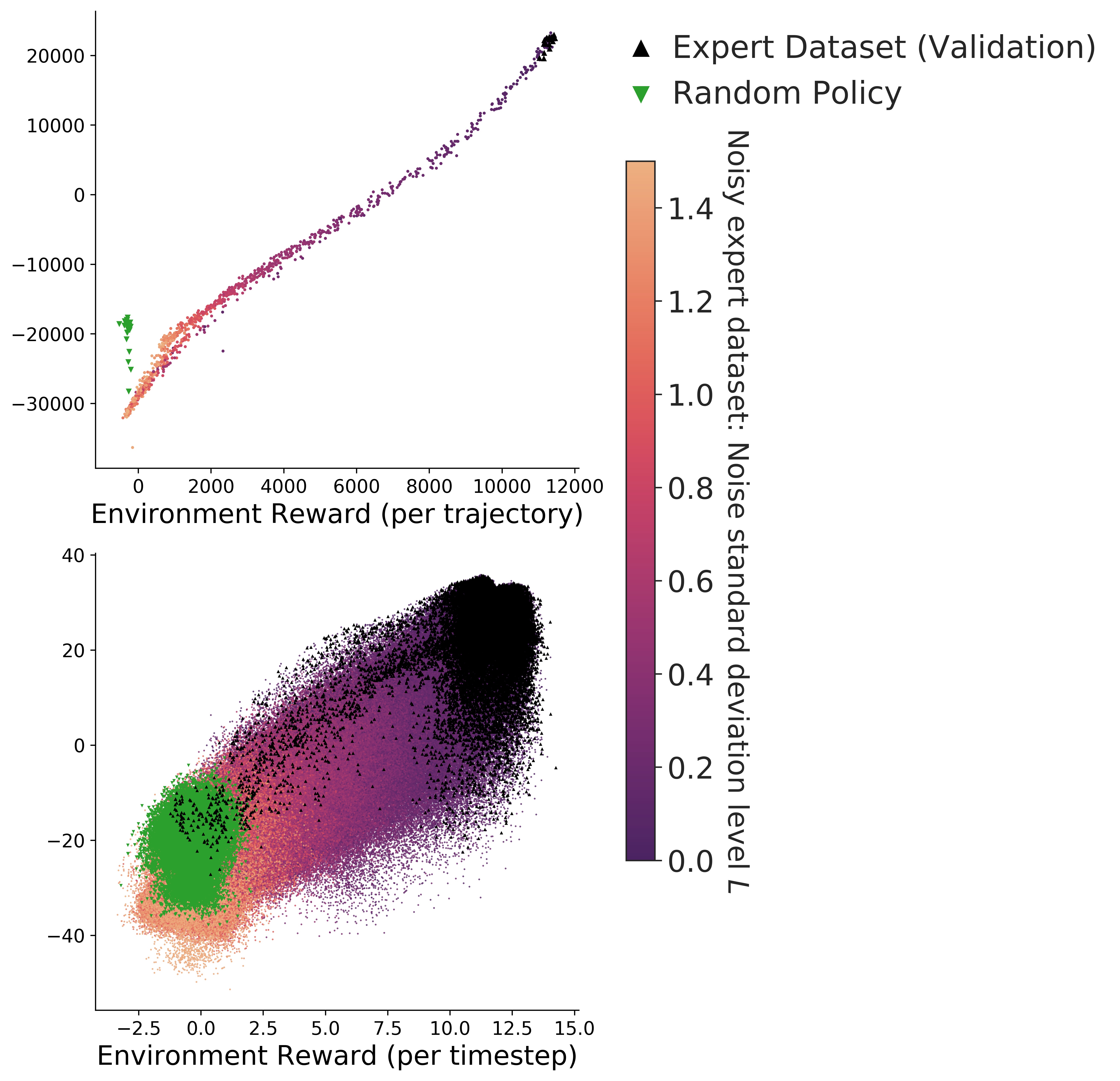}\\%{figures/10noises_dim_11_3.0.jpg}  \\
(a) Hopper & (b) Walker2d & (c)  HalfCheetah
\end{tabular}
\caption{Log probability as a function of environment rewards for Hopper, Walker2d, and HalfCheetah. Trajectory-wise (top row) and step-wise (bottom row). The expert demonstration validation dataset (black) has highest log probability (and environment reward), while a randomly initialized policy (green) gets assigned a very low log probability. See description of the noisy expert dataset in Appendix \ref{ap:dataset}.}
\label{fig:calibration}
\vspace{-4pt}
\end{figure*}

We report our results in Table \ref{table:results} and Figure \ref{fig:results}. IL-flOw outperforms all the baselines on all three studied environments, even with limited amounts of expert demonstrations. Notably it learns much faster than baseline algorithms. Figure \ref{fig:calibration} shows the relationship between the learnt reward signal and the environment reward. Our reward function is positively correlated with the environment reward and increases monotonically as we get closer to the expert behavior.
\vspace{-7pt}

\begin{table}[h]
\small
% \begin{center}
% \begin{tabular}{ccccc} 
% \toprule
%     {Method}  & {Hopper} & {Walker2d} & {HalfCheetah} \\ \midrule
%     {Expert Return}  &  $3420.40 \pm 33.97$ & $4370.09\pm 110.23$ & $11340.38 \pm 80.61$ \\ \midrule
%     {FKL (\textit{f}-IRL)~\cite{firl2020corl}}    & 3091.51 & 1663.67  & 7603.88  \\
%     {RKL (\textit{f}-IRL)~\cite{firl2020corl}}     & 3086.18 & 1369.33  & 7843.24 \\
%     {JS (\textit{f}-IRL)~\cite{firl2020corl}}     & 3161.76 &  1561.83 & 7931.70  \\ \midrule
%     {MaxEnt IRL~\cite{ziebart2008maximum}}   & 2376.16 & 1828.38 &  8197.89 \\ \midrule
%     {Our Method}    & \textbf{3312.64} & \textbf{4202.62}  &  \textbf{11710.86} \\
%  \bottomrule
% \end{tabular}
% \end{center}

\begin{center}
\begin{adjustbox}{center}
  \begin{tabular}{lccccccccc}
    \toprule
    % {Method}  & {Hopper} & {Walker2d} & {HalfCheetah} \\ \midrule
    % {Expert Return}  &  $3420.40 \pm 33.97$ & $4370.09\pm 110.23$ & $11340.38 \pm 80.61$ \\ \midrule
    
    Dataset &  \multicolumn{3}{c}{Hopper}{} &
      \multicolumn{3}{c}{Walker2d}{} &
      \multicolumn{3}{c}{HalfCheetah}{} \\
     Expert Return &  \multicolumn{3}{c}{$3420.40 \pm 33.97$}{} &
      \multicolumn{3}{c}{$4370.09\pm 110.23$}{} &
      \multicolumn{3}{c}{$11340.38 \pm 80.61$}{} \\
    \# Expert Traj.  & {10} & {20} & {40} & {10} & {20} & {40} & {10} & {20} & {40} \\
      \midrule
    {FKL (\textit{f}-IRL)
    % ~\citep{firl2020corl}
    } 
    & 3107.84 & 2772.57 & 3091.51 & 1811.41 & 2063.37 & 1663.67 & 8053.23 & 8432.35 & 7603.88\\
    {RKL (\textit{f}-IRL)
    % ~\citep{firl2020corl}
    } & 3187.05 & 3012.27 & 3086.18 & 1858.60 & 1519.23 & 1369.33 & 8039.80 & 8293.91 & 7843.24\\
    {JS (\textit{f}-IRL)
    % ~\citep{firl2020corl}
    } & 2459.27 & 3081.98 & \textbf{3161.76} & 1854.65 & 1844.41 & 1561.83 & \textbf{8123.40} & 8163.25 & 7931.70\\
    {MaxEnt IRL
    % ~\citep{ziebart2008maximum}
    } & 3171.23 & \textbf{3115.95} & 2376.16 & 1655.11 & 1787.43 & 1828.38 & 7853.19 & 8023.26 & 8197.89\\
    {Our Method}    & \textbf{3307.32} & \textbf{3139.89} &  \textbf{3312.64} & \textbf{4066.02} & \textbf{4254.20} & \textbf{4202.62} & 8043.43 & \textbf{11552.30} & \textbf{11710.86}\\
    \bottomrule
  \end{tabular}
  \end{adjustbox}
\end{center}

\caption{Final performance of different ILfO algorithms, using 10, 20, 40 expert demonstration trajectories, after 1M timesteps. All results are averaged across 3 seeds, with 10 evaluation rollouts per seed.}
 \label{table:results}
 \vspace{-7pt}
\end{table}

% \vspace{-10pt}
% \section{Conclusion}
% Although we have limited our scope to interactive settings, our full decoupling of the reward learning and policy learning approach allows for further applications in offline imitation.
% \scott{Do we even want a conclusion in a 4 page paper? Not sure how to end it though. I do think we we want this last paragraph in the introduction/elsewhere so the closing thought isn't "we're pretty similar to something else"}
% \greg{Say why using KL is preferred to Wasserstein.}
% \greg{And, never start a sentence with "And".}
% \weidi{Well tbh I actually think Wasserstein/EMD is a better smoother divergence metric more adapted to our case.}

% Use unnumbered third level headings for the acknowledgments. All acknowledgments
% go at the end of the paper. Do not include acknowledgments in the anonymized
% submission, only in the final paper.
% \newpage
% \section*{References}
\bibliographystyle{unsrt}
\bibliography{references}

\newpage
\appendix
\label{ap:dataset}
\section{Datasets Descriptions}
For visualization purposes we also collect for each environment a "noisy expert" dataset, which we'll refer to as $D_{Enoisy}$, consisting of 1000 trajectories from the expert policies with a fixed amount of additive Gaussian action noise sampled throughout each trajectory. Hence for a given trajectory in this dataset, we first sample a noise level $L \sim \mathcal{U}(0, 1.5)$, and we add to the expert actions at each timestep noise $\omega \sim \mathcal{N}(0, L)$. We find that a standard deviation $L = 1.5$ is enough to bring the expert policy to a random policy level of performance. This dataset allows us to inspect our learnt reward signal for trajectories in the environment generated by behaviors closer ($L\downarrow$) or further to the expert ($L\uparrow$) in policy space, the additive noise being on the actions. Further policies naturally translates to further state space explored in this noisy dataset as well. $D_{Enoisy}$ is shown in the `Flare' yellow to purple colormap on Figure \ref{fig:calibration}.

$D_{random}$ is the dataset containing transitions from the RL agent's warmup/exploration phase, using a uniform random policy $\pi_{rdm}(a | s) \sim \mathcal{U}(A_{min}, A_{max})$, with $A_{min}, A_{max}$ the environment action bounds. $D_{random}$ is shown in green on Figure \ref{fig:calibration}.

\section{Derivations for Eqns.~\ref{eq:entropies} and \ref{eq:approximation}}
\label{ap:equations}

Eq. \ref{eq:entropies}
\begin{align}
\mathcal{H}(p_\pi) &=  -\mathbb{E}_{s_{0:T-1} \sim p_{\pi}(\tau)}\left[\sum_{t=0}^{T-1} \log p_{\pi}(s_{t+1}|s_t)\right] \\
&= \sum_{t=0}^{T-1} -\mathbb{E}_{s_{0:T-1} \sim p_{\pi}(\tau)}\left[\log p_{\pi}(s_{t+1}|s_t)\right] \\
&= \sum_{t=0}^{T-1} -\mathbb{E}_{s_{0:t+1} \sim p_{\pi}(\tau)}\left[\log p_{\pi}(s_{t+1}|s_t)\right] \\
&= \sum_{t=0}^{T-1} \mathbb{E}_{s_{0:t} \sim p_{\pi}(\tau)}\left[ \mathcal{H}\left(p_{\pi}(\cdot|s_t)\right) \right] \\
&= \mathbb{E}_{s_{0:T} \sim p_{\pi}(\tau)}\left[\sum_{t=0}^{T-1} \mathcal{H}\left(p_{\pi}(\cdot|s_t)\right) \right].
\end{align}

Eq. \ref{eq:approximation}
\begin{align}
    \mathcal{H}\left(p_{\pi}(\cdot|s_t)\right) &= -\int p_{\pi}(s_{t+1}|s_t) \log p_{\pi}(s_{t+1}|s_t) \mathrm{d}s_{t+1} \\
    &= -\int \pi(a_t|s_t) \log \left(\left|\frac{\mathrm{d}a_{t}}{\mathrm{d}s_{t+1}}\right|\pi(a_t|s_t)\right) \mathrm{d}a_{t}\\
   &= -\int \pi(a_t|s_t) \log \pi(a_t|s_t) \mathrm{d}a_{t} - \int \pi(a_t|s_t) \log \left|\frac{\mathrm{d}a_{t}}{\mathrm{d}s_{t+1}}\right| \mathrm{d}a_{t} \\
   &= \mathcal{H}\left(\pi(\cdot|s_t)\right) + \mathbb{E}_{a_t \sim \pi}\left[\log \left|\frac{\mathrm{d}s_{t+1}}{\mathrm{d}a_{t}}\right| \right]
\end{align}
If the dynamics are approximately linear in the support of $\pi(a_t|s_t)$, then the second term becomes a constant and may be ignored for policy optimization. 
\section{Extra Experiments}
\label{ap:experiments}
% \subsection{Datasets}
% We collect $n=150$ demonstration trajectories using a Soft Actor Critic expert trained for 1M timestep, using $n$ random seeds and use a random subset of them as the training dataset $D_{expert}$ for the density estimator.

% For visualization purposes we also collect for each environment a "noisy expert" datasets, which we'll refer to as $D_{Enoisy}$, consisting of 1000 trajectories from the expert policies with a fixed amount of additive Gaussian action noise drawn for each trajectory. Hence for a given trajectory in this dataset, we first sample a noise level $L \sim \mathcal{U}(0, 2.5)$, and we add to the expert actions at each timestep noise $\omega \sim \mathcal{N}(0, L)$. We find that a standard deviation $L = 2.5$ is enough to bring the expert policy to a random policy level of performance. 

% Last, we'll refer to $D_{random}$ as the dataset containing transitions from the RL agent's warmup/exploration phase, using a uniform random policy $\pi_{rdm}(a | s) \sim \mathcal{U}(A_{min}, A_{max})$, with $A_{min}, A_{max}$ the environment action bounds.

\subsection{Noise Regularized Normalizing Flow}
We need to ensure this density model is representative of the expert demonstration state distributions. By using the generative capabilities of normalizing flows, which are the inverse operation of their density estimation mode, we can inspect the effect of the noise regularization process, as well as fit of our density model. 

Moreover, we should ensure that the expert dataset samples have the highest log likelihood
% \footnote{Or that a high environment reward gets assigned a high log probability if we are to assume the expert is optimal}
, and that samples away from the expert trajectories are assigned sensible log likelihoods. We can probe the trained transition density function anywhere and examine the sample's log probabilities.

\begin{figure*}
\centering
% Old version
% \begin{tabular}{cccc}
%   \includegraphics[width=.45\textwidth]{figures/w2d_dim4_0.0.jpg}&
%   \includegraphics[width=.45\textwidth]{figures/w2d_dim4_0.5.jpg}\\(a) h = 0.01 & (b) h = 0.5\\
%   \includegraphics[width=.45\textwidth]{figures/w2d_dim4_2.5.jpg}&
%   \includegraphics[width=.45\textwidth]{figures/w2d_dim4_4.5.jpg}\\ (c) h = 2.5 & (d) h=4.5
% \end{tabular}

\includegraphics[width=.65\textwidth]{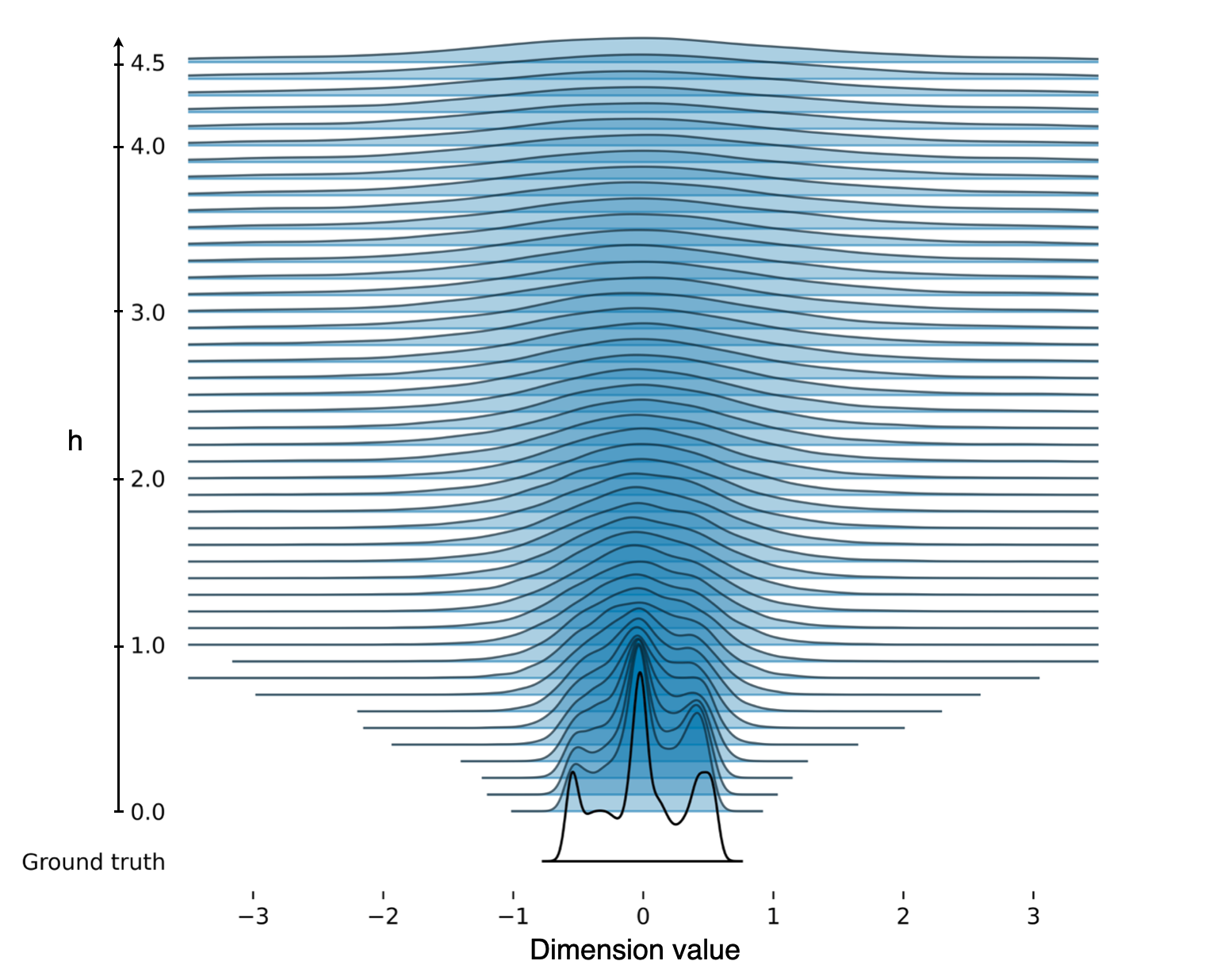}
\caption{$p_e(s' | s)$ density distributions for the 10th dimension of Hopper-v2, for the expert transition training dataset in black, and the normalizing flow model learnt noise conditional distribution in blue. Higher values of $h$ correspond to regularized versions of the training dataset distribution, smoothly extending its support.}
\label{fig:dims}
\end{figure*}

\paragraph{Single step sampling} : For each dataset $D_{E}, D_{Enoisy}$, given a sample $s$, we sample next state predictions $s'_{expert}$ given conditioning variables $s$ and a noise level $h \leq h_{max}$. We can visualize the noise regularization process this way. See Figure \ref{fig:dims}.\\

\paragraph{Reward calibration plots}: For each dataset $D_{E}, D_{Enoisy}, D_{random}$ we compare the environment ground truth reward with the learnt reward function. See Figure \ref{fig:calibration}.

\section{Hyperparameters}
\label{ap:hp}
\subsection{Soft Actor Critic}

\begin{table}[H]
\begin{center}
\begin{tabular}{cc} 
\toprule
    {Parameter}  & {Value}  \\ \midrule
    {Entropy regularization coefficient $\alpha$}     & 0.1  \\
    {Automatic entropy tuning}     & False  \\
    {$\tau$}     & 5e-4  \\
    {Actor network architecture (hidden)}     & [512, 512]  \\ 
    {Critic network architecture (hidden)}   &  [1024, 1024] \\ 
    {Actor LR}  &  3e-4  \\ 
    {Critic LR}    & 3e-4  \\
    {Optimizer}    & Adam  \\
    {Actor non linearity} & Tanh \\
    {Critic non linearity} & ReLU \\
 \bottomrule
\end{tabular}
\end{center}
\caption{SAC Hyperparameters}
 \label{table:sac_hyperparams}
\end{table}

\subsection{Neural Spline Flows}

\begin{table}[H]
\begin{center}
\begin{tabular}{cc} 
\toprule
    {Parameter}  & {Value}  \\ \midrule
    {Training epochs}  &  1000  \\ 
    {LR}  &  5e-4  \\ 
    {Spline bins}     & 8  \\
    {Network size (hidden)}     & [8, 8]  \\ 
    {Transform type}    & Rational quadratic coupling  \\ 
    {Mask type} & Alternating binary\\
    {Number of flow layers}     & 3  \\ 
    {Base distribution}    & Conditional Diagonal Normal  \\ 
    {Optimizer}    & AdamW  \\
    {Weight Decay} & 1e-4 \\
    {$h_{min}$} & 0.0 \\
    {$h_{max}$} & 4.5 \\
    {Non linearity} & Sine($\omega_0 = 2 \pi)$\\
    {Spectral Normalization} & True \\
 \bottomrule

\end{tabular}
\end{center}
\caption{Neural Spline Flow Hyperparameters}
 \label{table:nf_hyperparams}
\end{table}

\end{document}